\documentclass{article}
\usepackage{spconf,amsmath,graphicx}
\usepackage{float}

\title{AUGMENTING MOLECULAR IMAGES WITH VECTOR REPRESENTATIONS AS A FEATURIZATION TECHNIQUE FOR DRUG CLASSIFICATION}
\name{Daniel de Marchi, Amarjit Budhiraja}
\address{University of North Carolina at Chapel Hill, Dept. of Statistics and Operations Research}
\begin{document}
%
\maketitle
\begin{abstract}

One of the key steps in building deep learning systems for drug classification and generation is the choice of featurization for the molecules. Previous featurization methods have included molecular images, binary strings, graphs, and SMILES strings. This paper proposes the creation of molecular images "captioned" with binary vectors that encode information not contained in or easily understood from a molecular image alone. Specifically, we use Morgan fingerprints, which encode higher level structural information, and MACCS keys, which encode yes/no questions about a molecule's properties and structure. We tested our method on the HIV dataset published by the Pande lab, which consists of 41,127 molecules labeled by if they inhibit the HIV virus. Our final model achieved a state-of-the-art AUC-ROC on the HIV dataset, outperforming all other methods. Moreover, the model converged significantly faster than most other methods, requiring dramatically less computational power than unaugmented images.

\end{abstract}

\section{INTRODUCTION}
\label{sec:intro}

Deep learning for chemistry is a field still in its infancy, but with incredible potential. There are millions of potential drug molecules, yet we are only aware of around 12,000 such substances [1]. Data is accumulating rapidly, but is still often very sparse and difficult to work with compared to other problems in deep learning. Advances in both classifying and generating new molecules is showing incredible promise despite these challenges. As an example, Insilico Medicine published an article recently demonstrating an AI system that found several novel antibacterials in under a month [2]. 

We begin with a brief overview of previous molecular featurization methods to put our work in context. While there are a multitude of methods, only a few are relevant to this paper. The first is encoding the molecule as a SMILES string, which describes the atoms and bonds making up the molecule in a string format. It is a nonunique representation, and encodes the molecular graph by doing a depth-first traversal and marking each feature with an ASCII character [3]. The second set of techniques are binary vector representations, such as fingerprints and MACCS keys. 

Fingerprints denote the presence or absence of certain molecular substructures as a vector of binary variables, and are of arbitrary length. The procedure to generate these fingerprints is as follows. First, all atoms in a molecule are identified and assigned a value based on their properties and the structure of everything immediately connected to them. These values are then updated iteratively, expanding outwards to include detail about the molecular structure to a radius of two bonds away from the atom [4]. Larger radii can be used, but we kept to the default of two. This creates a set of which structures are present, from which duplicates are removed, and the result is hashed to a bit vector of a certain length, in our case 2,048. MACCS keys represent the answers to a fixed set of 167 yes/no questions about the molecule's structure and properties and represents that as a sequence of binary variables [5]. Finally, the Chemception images developed by Goh et al. use SMILES strings to reconstruct the molecular graph as a small image [6]. These are the featurizations we used to build our augmented images.

\begin{figure}[h]
    \centering
    \includegraphics[width=8cm]{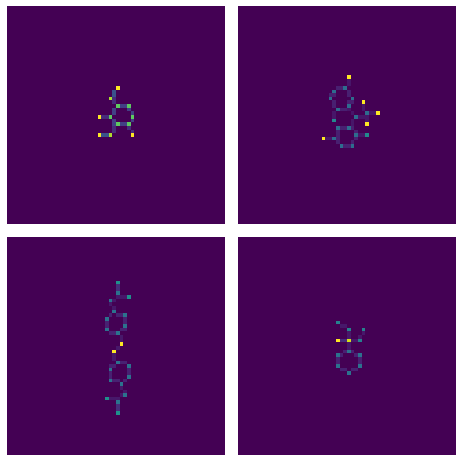}
    \caption{Examples of Chemception images}
\end{figure}

The choice of featurization method has been shown to have a large impact on the performance of both classifiers and generators of drug molecules. A common issue for generators is actually creating a valid molecule; SMILES is by far the most popular, but generators frequently produce invalid strings [7][8]. Classifiers based on fingerprints perform extremely well and the nature of the featurization helps prevent overfitting, but since it is a not a 1 to 1 mapping multiple molecules can have the same fingerprint, and molecules cannot be reconstructed from fingerprints. This makes fingerprints unsatisfactory in many use cases [9]. In short, each featurization has strengths and weaknesses, and performance can vary significantly from dataset to dataset and task to task.

The idea of creating augmented featurizations is not new. Goh et al. followed up their Chemception featurization with AugChemception, where they created multichannel images by pairing their original Chemception images with images containing information on partial charges, hybridization, and other values [10]. This work was later extended by Bjerrum et al. for several other image formats [11]. CheMixNet paired SMILES strings with fingerprints and MACCS keys, and demonstrated a significant improvement in AUC-ROC over any singular method [12]. While borrowing heavily from these ideas, our approach is unique to our knowledge and outperforms all other methods on this dataset. In addition, it is simple to implement and converges to a solution faster than any other method we are aware of.

\section{DATASET PREPARATION}
\label{sec:format}

The HIV dataset contains 41,127 molecules whose activity on the HIV virus is known, encoded as SMILES strings [13]. We generated all three representations (Chemception images, MACCS keys, and fingerprints), setting the size limit of the image to be 60x60. While this did exclude some molecules, we overall did not reduce the size of the dataset by a significant amount. To include every image we would have to use 160x160 images, which would dramatically increase training time and make the resulting images very sparse, so we elected to use smaller images. 

There is one limitation of the HIV dataset, which is that only 1,443 of the 41,127 molecules inhibit the virus. We were left with 1,258 of these molecules after removing the ones that produced very large images, which also slightly reduced our number of non-inhibitory molecules [6]. The Chemception paper chose to upsample the minority class, as did the majority of other papers using this dataset, so we elected to take this approach as well. We did not elect to normalize the images, as we replicated the performance in the Chemception paper without any additional modifications.

\section{NEURAL NETWORK DESIGN}
\label{sec:pagestyle}

The Chemception paper used a modified version of Google's Inception-Resnet architecture [6][14]. To mimic the original paper and provide a fair comparison, we replicated their approach as faithfully as was possible from their description. A lengthier description of their methods is available in their paper, and we will not be going into everything they chose to do here. They used three types of blocks similar to the Inception-Resnet v2 blocks, with their best performing model having three of each type of block in sequence, and a filter size of 16 on the convolutional layers [14]. We actually found improvement up to four of each block, contrary to their paper [6]. This is likely because their implementation differs slightly from ours, and our reconstruction is inexact. We also had to remove the final convolutional layer on the A, B, and C blocks due to overfitting issues on our more complex featurizations, further contributing to small differences in outcomes. 

\begin{figure}[h]
    \centering
    \includegraphics[width=8.5cm]{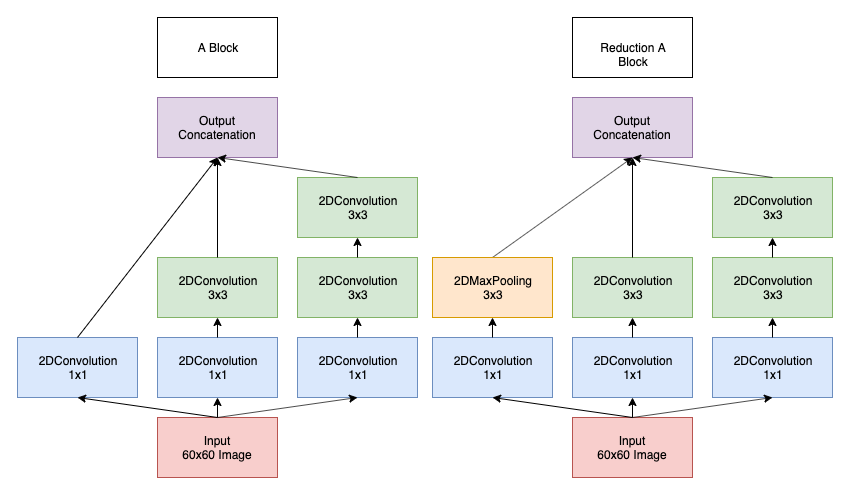}
    \caption{Our A Block and A Reduction Block architectures}
\end{figure}

We used a learning rate of 0.001, a batch size of 32, and the Adam optimizer with learning rate decay. Specifically, we set the patience to five, with the learning rate halving if there was no improvement for five epochs. We altered the training batches by introducing rotations and horizontal and vertical translations to the images, helping to prevent overfitting. 

\begin{figure}[h]
    \centering
    \includegraphics[width=8.5cm]{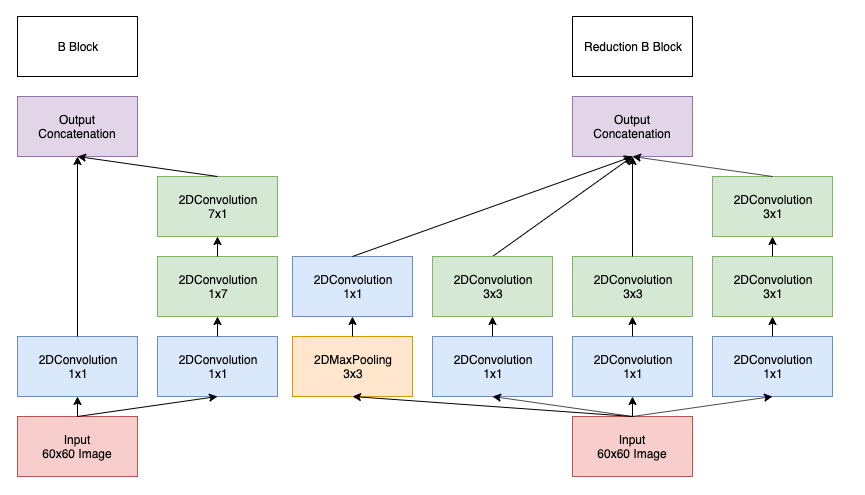}
    \caption{Our B Block and B Reduction Block architectures}
\end{figure}

We used a dense layer with a single neuron to handle the fingerprints, and two dense layers, one with five neurons and an output with a single neuron, to handle the MACCS. We observed that anything more complex resulted in dramatic overfitting within a couple of epochs, which is very different from the CheMixNet paper where the additional representations were handled with a network almost as complex as the one handling the primary SMILES representation [12]. We treated MACCS differently because the AUC-ROC improved slightly for a more complex network before overfitting. This is possibly due to the higher quality of Chemception images as a representation, or the fact that the CheMixNet authors chose to downsample the molecules that had no effect on HIV rather than upsample the minority class [6][12]. 

\begin{figure}[h]
    \centering
    \includegraphics[width=8.5cm]{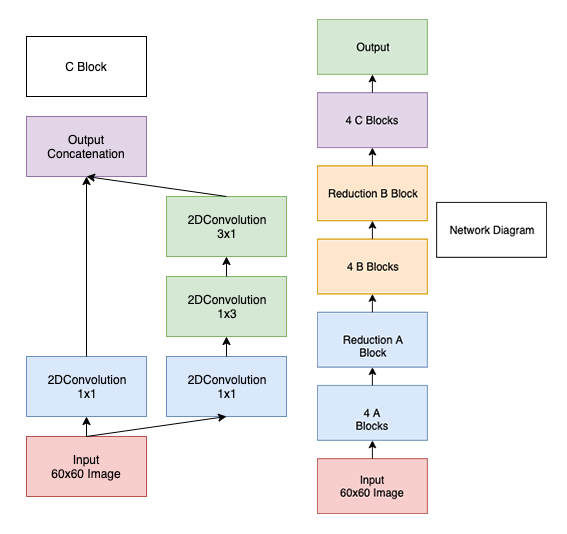}
    \caption{Our C Block and overall network architectures}
\end{figure}

\section{RESULTS}
\label{sec:typestyle}

The image only model is very comparable to the results in the Chemception paper [6]. Due to limited computational resources, the extensive training required for this featurization, and the fact that we were simply repeating an earlier result, we did not apply 5-fold cross validation, instead only testing the model once on 20\% of the dataset. Our training protocols are fairly similar, and the ROC-AUC achieved is within 1\% of the ROC-AUC achieved in the Chemception paper, so we assumed that this was a sufficiently faithful replication and within a normal margin of error for a repeated experiment. The ROC-AUC achieved on the 20\% test set was 0.749, compared to 0.748 from Chemception. For all further models, we applied 5-fold cross validation.

The first featurization method tested was to augment the image with MACCS keys. This provided a modest improvement, increasing ROC-AUC from 0.748 to 0.7733, with a range across the folds of 0.7625 to 0.7852. 

The images paired with fingerprints was more successful. Fingerprints are a much more complex representation, and have no standard length in contrast to MACCS keys. Previous work has indicated that using larger fingerprints produced better results, so we elected to use fingerprints of length 2,048 rather than the standard 1,024 [15]. This this took the average ROC-AUC to 0.7955 with a range from 0.7847 to 0.8085.

Our final step was to test all three featurizations together. This had a synergistic effect, improving images alone by more than either fingerprints or MACCS had alone. This achieved a mean ROC-AUC of 0.8567, with a range from 0.8351 to 0.8751. 

\begin{figure}[H]
    \centering
    \includegraphics[width=8cm]{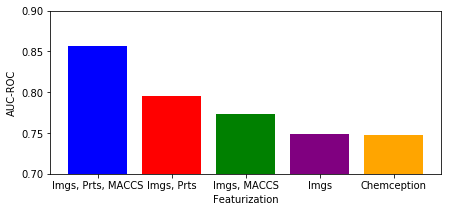}
    \caption{Comparison of various captioned image options, with imgs being Chemception images, prts being fingerprints, and MACCS being MACCS keys}
\end{figure}

Finally, adding in additional featurizations simplified the training protocol significantly. The plot below summarizes both the time per epoch and the number of epochs to achieve the best solution. 

\begin{figure}[H]
    \centering
    \includegraphics[width=8cm]{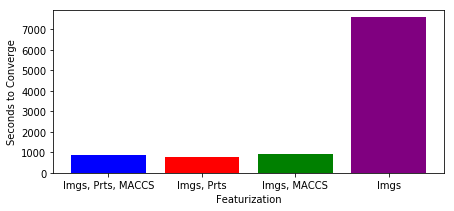}
    \caption{Seconds to converge for each featurization, running on a GTX 1080}
\end{figure}

We can also compare our final representation with all three featurizations to the other state of the art techniques. We elected not to include CheMixNet's results due to their sampling procedure, which differed from the procedure most papers used, and significantly affected their results. Their implementation of Chemception images achieved an AUC-ROC close to 0.85 despite no changes to the method, indicating that downsampled results are fundamentally not comparable to upsampled results [12]. 

\begin{figure}[h]
    \centering
    \includegraphics[width=8cm]{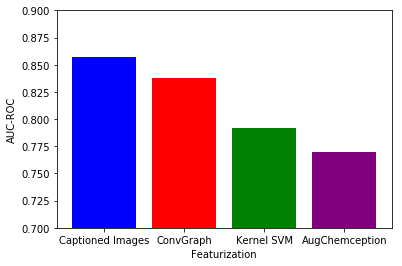}
    \caption{Captioned images vs. other high performing methods [13][10][16]}
\end{figure}

\section{DISCUSSION}
\label{sec:majhead}

Overall, these results are quite promising. Augmenting the Chemception images with binary vectors provided significant benefits, and performed better than adding additional image "channels" or representing the molecular graph differently as was attempted in other papers. It also simplified and shortened the training procedure considerably, which depending on the size of the dataset in question could be almost as valuable as the improvement to the classifier. 

It is likely that the additional representations helped for two reasons. The MACCS keys provide chemistry information that is not discernible from a molecular graph or image, and some of that information appears to have an impact on a molecule's ability to inhibit the HIV virus. The fingerprints provide a higher level view of a molecule by looking at substructures rather than individual atoms and bonds.

There are some noticeable limitations to this study however. First, only one dataset was used. While this was an effective dataset as a proof of concept, it does not provide a good indication of the power of this technique across the full range of possible datasets. Both Chemception and AugChemception had their relative performance change significantly depending on the task [6][10]. Furthermore, the simplified, faster training protocol could be seen as a negative. It is almost certain that with some work to prevent overfitting, some small gains to the ROC-AUC could be had at the expense of significantly more computing power and time. Depending on use case, this could be more valuable. Ultimately, we elected not to pursue this because to prevent overfitting our image classifier would have to be significantly simplified or altered, and that would provide a less accurate comparison between this paper and the Chemception work. Lastly, one of the goals of Chemception was to be domain agnostic and not make too many assumptions about the information needed for the problem. While we feel using fingerprints remains true to this vision as they contain no information beyond the molecular structure, MACCS keys do not. However, most problems are not domain agnostic, and MACCS keys are used very broadly across all subfields of chemistry, so we feel that relaxing the original intentions of Chemception images is warranted.

This work does suggest some interesting avenues to pursue in future papers. An example would be to explore additional image/vector pairs, such as by using the multi channel images from AugChemception, or to pair images with SMILES strings. It seems that molecular graphs alone, no matter the representation, are not fully optimal for classification problems. Moreover, the choice of augmentation has a strong effect on performance, and using more varied representations may be better than an array of similar representations.

\section{CONCLUSION}

In this work, our goal was to come up with a novel augmentation to Chemception images that was better than previous attempts. By adding binary vectors that encoded properties that cannot be determined from a molecular graph, we created a classifier that is faster to train, easy to implement, and has a state-of-the-art AUC-ROC. 

\bibliographystyle{IEEEbib}
\bibliography{strings,refs}

\section{BIBLIOGRAPHY}

[1] Wishart et al., “DrugBank 5.0: a major update to the DrugBank database for 2018,” Nucleic Acids Research, vol. 46, pp. D1074-D1082, 2017.

[2] Ivanenkov et al., “Identification of Novel Antibacterials Using Machine Learning Techniques,” Front. Pharmacol., doi:10.3389/fphar.2019.00913, 2019.

[3] D Weininger, “SMILES, a chemical language and information system. 1. Introduction to methodology and encoding rules,” Journal of Chemical Information and Modeling, vol. 28, pp. 31-36, 1988.

[4] D Rogers, M Hahn, “Extended-Connectivity Fingerprints,” J. Chem. Inf. Model., vol. 50, pp. 742-754, 2010.

[5] Durant et al., “Reoptimization of MDL Keys for Use in Drug Discovery,” J. Chem. Inf. Comput. Sci., vol. 42, pp. 1273-1280, 2002.

[6] Goh et al., “Chemception: A Deep Neural Network with Minimal Chemistry Knowledge Matches the Performance of Expert-developed QSAR/QSPR Models,” arXiv:1706.06689, 2017.

[7] M Olivecrona, T Blaschke, O Engkvist, H Chen, “Molecular de-novo design through deep reinforcement learning,” Journal of Cheminformatics, vol. 9, article 48, 2017.

[8] M Kusner, B Paige, J Hernandez-Lobato, “Grammar Variational Autoencoder,” ICML, 2017.

[9] D Elton, M Fuge, P Chung, “Deep learning for molecular design—a review of the state of the art,” Molecular Systems Design and Engineering, vol. 4, 2019.

[10] Goh et al., “How Much Chemistry Does a Deep Neural Network Need to Know to Make Accurate Predictions?,” IEEE Winter Conference on Applications of Computer Vision, 2018.

[11] E Bjerrum, B Sattarov, “ Improving Chemical Autoencoder Latent Space and Molecular De novo Generation Diversity with Heteroencoders,” Biomolecules, vol. 8, pp 131, 2018.

[12] A Paul et al., “CheMixNet: Mixed DNN Architectures for Predicting Chemical Properties using Multiple Molecular Representations,” NIPS, 2018.

[13] Wu et al., “MoleculeNet: A Benchmark for Molecular Machine Learning,” Chemical Science, vol. 2, 2018.

[14] C Szegedy, S Ioffe, V Vanhoucke, A Alemi, “Inception-v4, Inception-ResNet and the Impact of Residual Connections on Learning,” AAAI, 2017.

[15] N O'Boyle, R Sayle, “Comparing structural fingerprints using a literature-based similarity benchmark,” J Cheminform, vol. 8, Article 36, 2016.

[16] S Kearnes et al., “Molecular Graph Convolutions: Moving Beyond Fingerprints,” Journal of Computer-Aided Molecular Design, vol. 30, pp. 595-608, 2016.

\end{document}